\documentclass[letterpaper, 10pt, conference]{ieeeconf}
\IEEEoverridecommandlockouts
\makeatletter
\let\NAT@parse\undefined
\makeatother

\usepackage[numbers, sort&compress]{natbib}
\usepackage[dvipsnames,table]{xcolor}
\usepackage{times}
\usepackage{amsmath}
\usepackage{amssymb}
\usepackage{bm}
\usepackage{graphicx}
\usepackage{booktabs}
\usepackage{multicol}
\usepackage{multirow}
\usepackage{makecell}
\usepackage{adjustbox}
\usepackage{array}
\usepackage{subcaption}
\usepackage{stfloats}
\usepackage{float}
\usepackage{arydshln}
\usepackage{etoolbox}
\usepackage{pifont}
\usepackage{tcolorbox}
\usepackage{tikz}
\usepackage{caption}
\usepackage{xspace}

\definecolor{V}{HTML}{B2E4B2}
\definecolor{S}{HTML}{deb69b}
\definecolor{B}{HTML}{D1E5FD}
\definecolor{VB}{HTML}{88e3e3}
\definecolor{SB}{HTML}{E3C7FF}
\definecolor{VSB}{HTML}{EAEAEA}
\definecolor{control_yellow}{HTML}{dbd400}
\definecolor{lightblue}{HTML}{5590ed}
\definecolor{darkgreen}{rgb}{0.0, 0.5, 0.0}

\usepackage{pdfx}
\hypersetup{
    colorlinks=true,
    linkcolor=lightblue,
    filecolor=lightblue,
    urlcolor=lightblue,
    citecolor=lightblue,
}
\usepackage[capitalize]{cleveref}

\newcommand{\mypar}[1]{\vspace{1em}\noindent\textbf{#1}}

\title{\LARGE \bf REALM: A Real-to-Sim Validated Benchmark for Generalization in Robotic Manipulation}

\author{
    Martin Sedlacek$^{1,2,\ast,\dagger}$ \quad
    Pavlo Yefanov$^{1,2,\ast}$ \quad
	Georgy Ponimatkin$^{1,2}$ \quad
    Jai Bardhan$^{1}$ \quad
    Simon Pilc$^{1,2}$ \\
	Mederic Fourmy$^{1}$ \quad
	Evangelos Kazakos$^{1}$ \quad
	Cees G. M. Snoek$^{3}$ \quad
	Josef Sivic$^{1}$ \quad
	Vladimir Petrik$^{1}$ \\
    $ $ \\
    $^{1}$Czech Institute of Informatics, Robotics and Cybernetics, Czech Technical University in Prague \\
    $^{2}$Faculty of Electrical Engineering, Czech Technical University in Prague \quad
    $^{3}$University of Amsterdam
    \thanks{$^*$Equal contribution.}
    \thanks{$\dagger$Work done in part as a student at the UvA visiting CIIRC.}
}

\IEEEaftertitletext{
    \begin{center}
    \includegraphics[width=\textwidth]{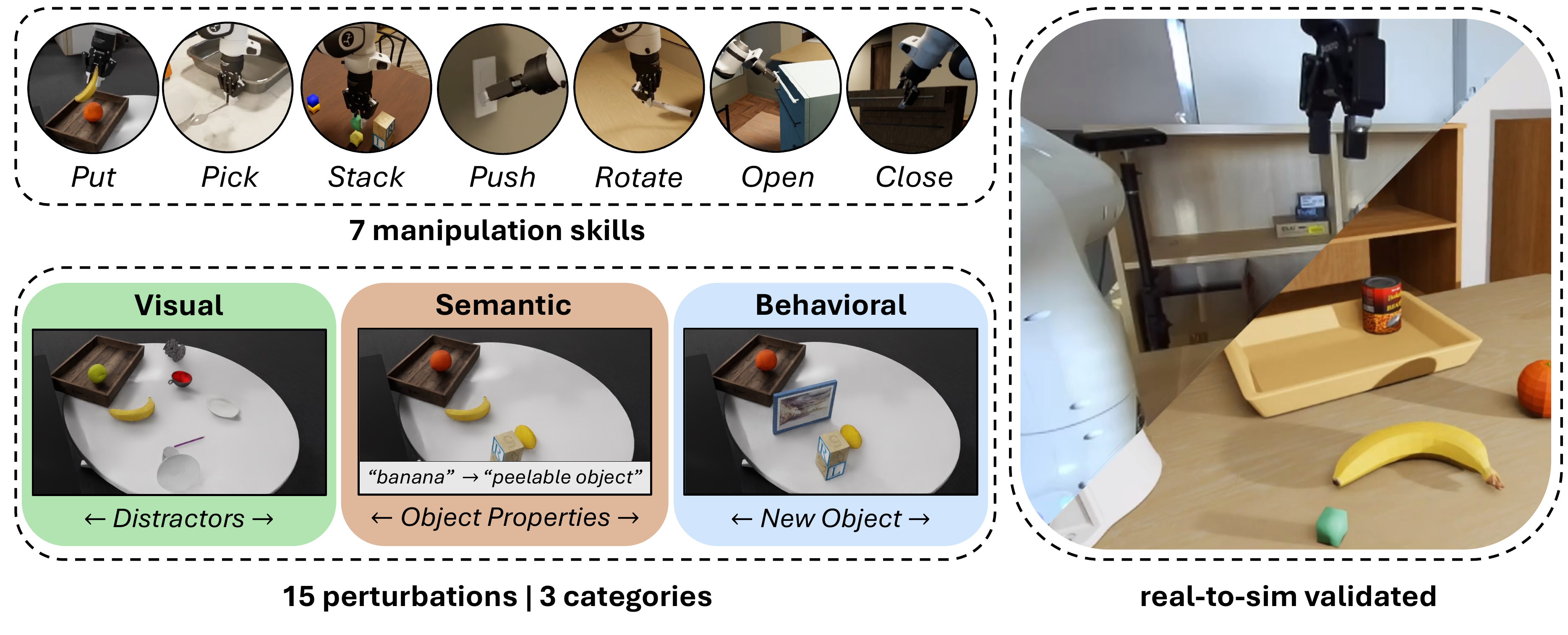}
    \captionof{figure}{\textbf{REALM} is a large-scale real-to-sim aligned simulation environment and benchmark for generalization in robotic manipulation. It supports 7 distinct manipulation skills (top left) and stress-tests them against 15 perturbations (bottom left). Through empirical validation, we show that evaluation results in simulation are strongly correlated to real-world performance.
    }
    \label{fig:teaser}
    \end{center}
    \vspace{1em}
}

\begin{document}

\maketitle
\thispagestyle{empty}
\pagestyle{empty}

\begin{abstract}
Vision-Language-Action (VLA) models empower robots to understand and execute tasks described by natural language instructions. However, a key challenge lies in their ability to generalize beyond the specific environments and conditions they were trained on, which is presently difficult and expensive to evaluate in the real-world. To address this gap, we present REALM, a new simulation environment and benchmark designed to evaluate the generalization capabilities of VLA models, with a specific emphasis on establishing a strong correlation between simulated and real-world performance through high-fidelity visuals and aligned robot control. Our environment offers a suite of 15 perturbation factors, 7 manipulation skills, and more than 3,500 objects. Finally, we establish two task sets that form our benchmark and evaluate the $\mathbf{\pi_0}$, $\mathbf{\pi_0}$-FAST, and GR00T N1.5 VLA models, showing that generalization and robustness remain an open challenge. More broadly, we also show that simulation gives us a valuable proxy for the real-world and allows us to systematically probe for and quantify the weaknesses and failure modes of VLAs. Project page: \url{https://martin-sedlacek.com/realm/}
\end{abstract}
\section{Introduction}

Reproducible benchmarks and standardized evaluation protocols have been crucial for accelerating progress in numerous fields, including natural language processing~\cite{wang2019gluemultitaskbenchmarkanalysis, hendryck2021mmlu},
computer vision~\cite{imagenet, lin2015mscoco, nguyen2025bop24, thoker2022severe_benchmark}, and computational biology~\cite{kryshtafovych2023critical, Jumper2021AlphaFold}.
In the rapidly evolving field of Vision-Language-Action (VLA) models~\cite{black2024pi0, pertsch2025fastefficientactiontokenization, nvidia2025gr00tn1openfoundation, brohan2023rt2, geminiroboticsteam2025geminirobotics, kim2024openvla}, which aims to equip robots with general manipulation skills and understanding of instructions expressed in natural language, the need for robust evaluation methods is even more paramount, given their ability to directly interact with the physical world. While the training datasets typically span multiple robots, environments, tasks, and objects, the model evaluation often occurs on a non-standardized real-world set-up, posing significant challenges in terms of reproducibility and cost, leading to limitations in the scale of robotic evaluation. This is especially problematic for evaluating generalization capabilities, which requires diversity across objects, scenes, tasks, and large sample sizes.

An alternative to real-world evaluation~\cite{atreya2025roboarena, gao2025generalist_taxonomy} is physical simulation, offering the potential for controlled experiments and cost-effective evaluation. However, existing simulation benchmarks for robotic manipulation typically focus on a limited variety of perturbations~\cite{garcia2025gembench, zhang2024vlabench, pumacay2024colosseum, li24simpler} and lack high-fidelity visuals and realistic control. This is a critical shortcoming when evaluating policies against the complexities of real-world dynamics. 

To address these limitations, our work offers three main contributions. First, we present a reproducible, high-fidelity simulation environment with aligned robot control. We implement 7 skills, 15 controlled perturbations, and support thousands of objects, enabling reliable, large-scale evaluation of generalization capabilities in VLAs. Second, we establish that our high-fidelity simulation serves as a good proxy for real-world performance through extensive real-to-sim validation on nearly 800 pairs of real and simulated rollouts across multiple individual perturbation factors, showing a strong correlation between model performance in simulation and the real world. Third, we propose a new generalization benchmark inside our simulation environment and evaluate three state-of-the-art VLA models. We observe that generalization across most perturbation factors, as well as robustly completing the tasks is still highly challenging, despite the models being trained on large-scale robot data. Many of our findings are further supported by previously observed failure modes from smaller scale real-world testing~\cite{gao2025generalist_taxonomy, Wang2025pi_in_the_wild}.
\section{Related work}

The recent progress in robot learning has shown that current simulation benchmarks~\cite{liu2023libero, gong2023arnold} and evaluation protocols can be quickly saturated~\cite{reuss2025state_of_vla_research_iclr26} and often fail to generate a meaningful signal about real-world model performance. In this section, we discuss the existing benchmarks and the need for large-scale reproducible evaluation in the era of VLAs.

\mypar{Evaluating generalization in robotic manipulation.}
Nearly all benchmarks for generalization in robotic manipulation~\cite{li2024behavior1k, li24simpler, xing2021kitchenshift, garcia2025gembench, zheng2022vlmbench,liu2023libero, pumacay2024colosseum, gong2023arnold, zhang2024vlabench, mees2022calvinbenchmarklanguageconditionedpolicy} are simulation-based due to the reproducibility requirement and the high desired variety of objects, scenes, and perturbations. Evaluating generalization in a real-world setting, even with relatively small sample sizes, still requires hundreds of rollouts \cite{gao2025generalist_taxonomy}, quickly becoming unsustainable and difficult to reproduce adequately. Efforts such as RoboArena~\cite{atreya2025roboarena} are a great step towards scaling real-world evaluations with a distributed network of evaluation stations that run blind A/B evaluations of remotely hosted policies, but this approach does not systematically probe generalization across individual perturbation factors and the set-ups can change frequently, which hinders reproducibility. 
The rigorous tests on the effect of data scaling and pre-training for Large Behavior Models (LBMs)~\cite{trilbmteam2025carefulexaminationlargebehavior} similarly required hundreds of reproducible evaluations, relying heavily on high-fidelity simulation to obtain a statistically significant signal about policy performance.

\begin{table}[t!]
    \centering
    \vspace{1mm}
    \caption{Comparison of generalization benchmarks for robotic manipulation. The ``Perturbations'' column shows the number of considered \textbf{V}isual, \textbf{S}emantic, and \textbf{B}ehavioral perturbations as \emph{V/S/B}. The middle column indicates support for \textbf{H}igh-fidelity \textbf{V}isuals (\textbf{HV}), real-to-sim \textbf{A}ligned \textbf{C}ontrol (\textbf{AC}), and \textbf{M}ultiple \textbf{V}iews (\textbf{MV}). The ``Diversity'' column shows the number of supported \textbf{S}kills, S\textbf{c}enes, and \textbf{O}bjects as \textit{S/C/O}. REALM supports the most perturbations, especially in the behavioral category, as well as the largest variety of objects, alongside high-fidelity visuals and aligned control.}
    \label{table:benchmark_comparison}
    \begin{tabular}{>{\centering\arraybackslash}p{2cm}>{\centering\arraybackslash}p{1.5cm}>
    {\centering\arraybackslash}p{1.5cm}>
    {\centering\arraybackslash}p{2cm}}
    \toprule
    \textbf{Benchmark} &
    \textbf{\centering Perturbations (V/S/B)} &
    \textbf{\centering HV/AC/MV} &
    \textbf{\centering Diversity (S/C/O)} \\
    \midrule
    GemBench~\cite{garcia2025gembench}   & 2 / 1 / 2 & \textcolor{red}{\ding{55}} / \textcolor{red}{\ding{55}} / \textcolor{darkgreen}{\textbf{\ding{51}}} & 7 / 1 / 50+ \\
    VLABench~\cite{zhang2024vlabench}   & 1 / 7 / 2 & \textcolor{red}{\ding{55}} / \textcolor{red}{\ding{55}} / \textcolor{darkgreen}{\textbf{\ding{51}}} & \textbf{10} / \textbf{20} / 2,000+ \\
    COLOSSEUM\cite{pumacay2024colosseum}  & 5 / 0 / 2 & \textcolor{red}{\ding{55}} / \textcolor{red}{\ding{55}} / \textcolor{darkgreen}{\textbf{\ding{51}}} & \textbf{10} / 1 / 20+ \\
    SIMPLER~\cite{li24simpler} & 5 / 0 / 2 & \textcolor{darkgreen}{\textbf{\ding{51}}} / \textcolor{darkgreen}{\textbf{\ding{51}}} / \textcolor{red}{\ding{55}} & 5 / 3 / 10+ \\
    \midrule
    REALM (ours) & \textbf{6 / 8 / 7} & \textcolor{darkgreen}{\textbf{\ding{51}}} / \textcolor{darkgreen}{\textbf{\ding{51}}} / \textcolor{darkgreen}{\textbf{\ding{51}}} & 7 / 10 / \textbf{3,500+} \\
    \bottomrule
    \end{tabular}
\end{table}

\mypar{The real-to-sim gap}. Simulated evaluation typically requires co-training on in-domain data due to the inherent real-to-sim gap characterized by two distinct challenges~\cite{li24simpler}: \textbf{(1)}~visual fidelity and \textbf{(2)}~control alignment. Many widely used simulation frameworks lack high-fidelity visuals, causing a distribution shift that pushes vision-based policies trained on real data out of distribution, in turn artificially deteriorating their performance. Although high-fidelity simulators \cite{nvidia2025isaaclabgpuacceleratedsimulation, li2024behavior1k} can mitigate the visual gap, the underlying control is often misaligned, causing the resulting states from executing the same actions in simulation and reality to diverge. 

Mitigating the real-to-sim gap during policy evaluation is a known problem~\cite{li24simpler, abouchakra2025realissim, tao2024maniskill3}. For example, SureSim~\cite{badithela2025suresim} uses a small number of real evaluations and augments them with a scalable simulation to obtain reliable results. Automated real-to-sim reconstruction~\cite{jangir2025robotarenainftyscalablerobot, geng2025roboverse, reuss2025state_of_vla_research_iclr26, dai2024acdc, torne2024rialto} of environments in simulation is also being actively explored for both evaluation and training. To the best of our knowledge, reliable fully simulated evaluation of policies trained on real data is currently only practical and established in the SIMPLER benchmark~\cite{li24simpler}. 
While this is an exciting step, SIMPLER considers only a limited number of robot skills and objects, and only supports a single viewpoint. We address the limitations of current simulated benchmarks by using high-fidelity simulation to mitigate the visual gap and optimize robot physics to achieve realistic control, while offering a wide variety of skills, scenes, and objects. We further discuss the validation of our simulation in Section~\ref{section:real2sim} and provide a comparison with existing generalization benchmarks in Table~\ref{table:benchmark_comparison}.

\mypar{Vision-Language-Action (VLA) models}.
Driven by the success of scaling~\cite{kaplan2020scalinglawsneurallanguage} observed in vision and language, early works such as RT-1 \cite{brohan2023rt1} trained multi-task visuomotor policies as an end-to-end neural network using a large-scale dataset. Since then, VLAs which build upon Internet-scale pre-trained VLMs~\cite{beyer2024paligemmaversatile3bvlm, karamcheti2024prismatic} have started to show generalization across scenes and objects~\cite{sapkota2025vlasurvey}, mainly thanks to training on large-scale robot datasets~\cite{walke2024bridgedatav2, oxe2024oxe, khazatsky2025droid}.

\mypar{World Models}. Recent advances in action conditioned world models~\cite{guo2025ctrlworld} have shown that they can successfully be trained on a diverse dataset such as DROID. However, their primary shortcoming for benchmarking generalization is the lack of granular control over specific parameters such as object mass, pose, or scene illumination. The explicit guarantees about physics provided by traditional simulators are also inherently not present in learned world models. 
\begin{figure}[t!]
    \centering
    \vspace{3mm}
    \includegraphics[width=\columnwidth]{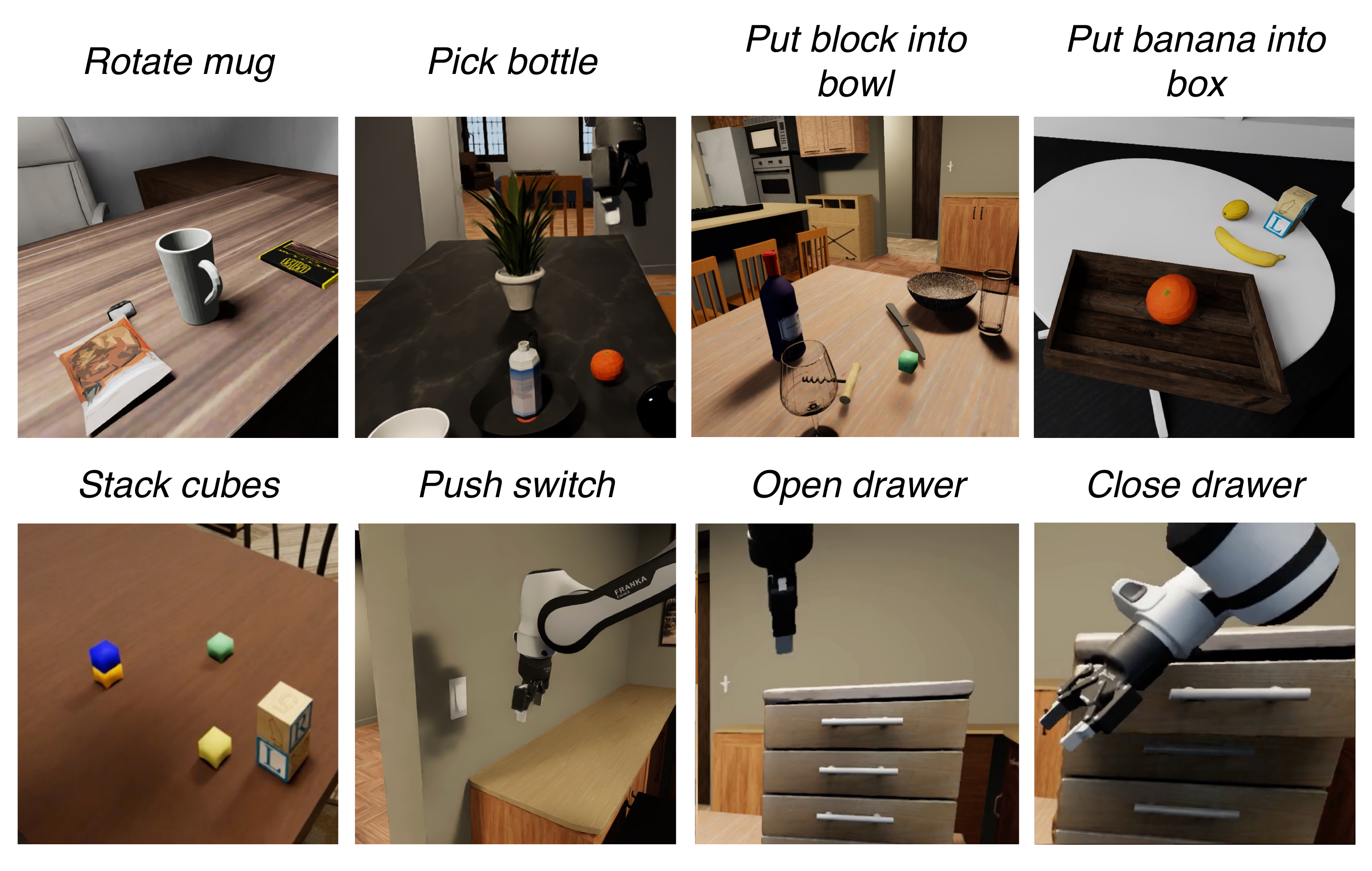}
    \caption{\textbf{Visualization of tasks from \textit{REALM-base} and \textit{REALM-articulated}}. The full benchmark consists of 8 base tasks (6 shown) and 2 articulated tasks (both shown), which are used for the large-scale evaluation in Section~\ref{section:Results}.}
    \label{fig:sim_gen_vis}
\end{figure}

\section{REALM: Generalization Benchmark}
Our benchmark is inspired by the DROID~\cite{khazatsky2025droid} platform and designed for evaluating generalization capabilities of VLAs and robotic foundation models without the need to explicitly co-train on simulated data. As seen in Table~\ref{table:benchmark_comparison}, the combination of realism, systematic perturbations, and diversity then allows us to establish a trusted evaluation protocol and probe for failure modes of the tested models. In this section, we cover the design choices of REALM, focusing primarily on the choice of perturbations, skills, and metrics, followed by details on minimizing the control gap. 
\begin{figure}[t!]
    \centering
    \vspace{3mm}
    \includegraphics[width=\columnwidth]{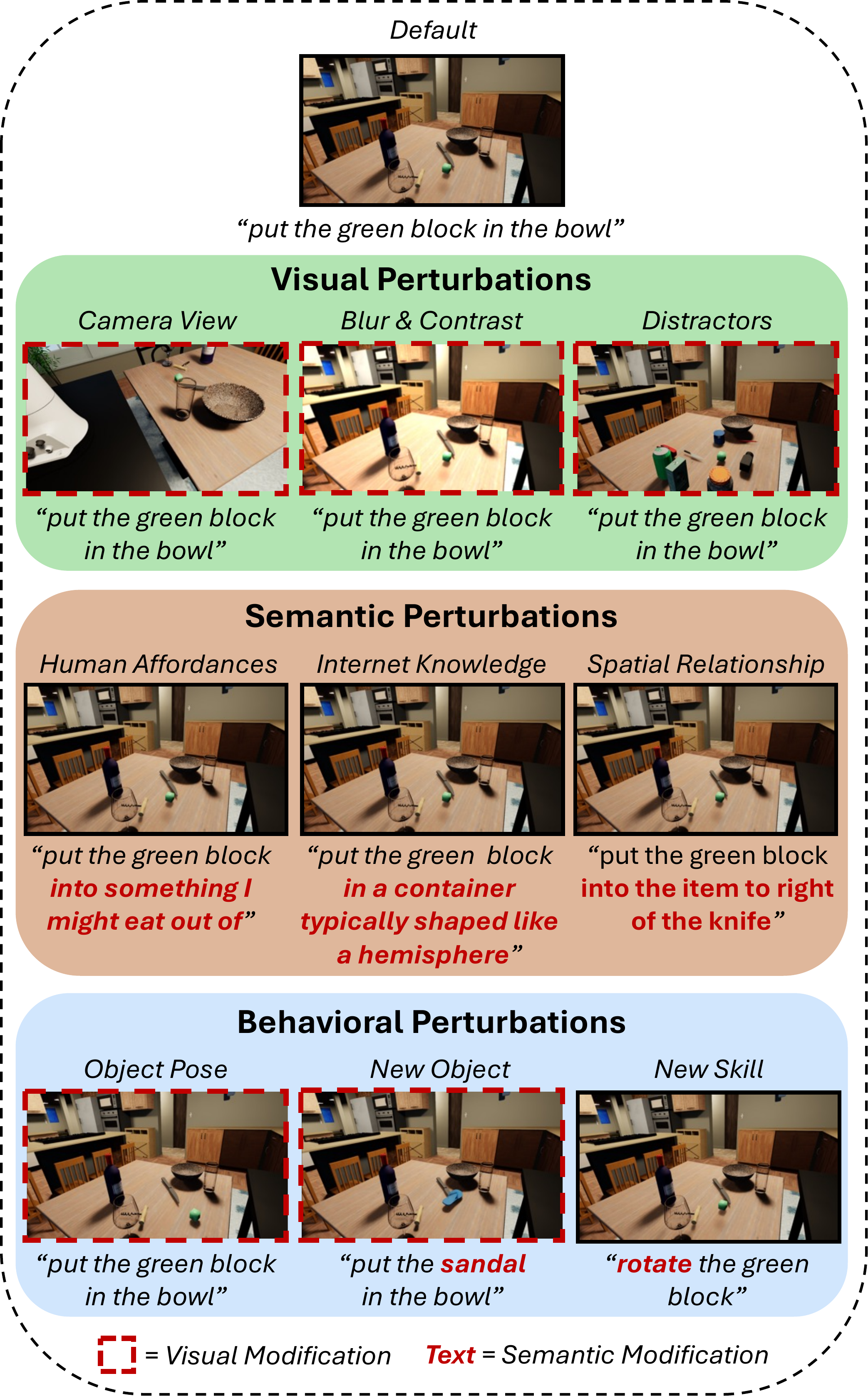}
    \caption{\textbf{Visualization of 9 out of 15 perturbations from three categories}.
    \textit{\textbf{V}isual} perturbations alter the observations in pixel space, but do not require the policy to adjust its behavior or understand a different phrasing of an instruction. \textit{\textbf{S}emantic} perturbations require understanding different aspects of human language that can be present in robot commands. \textit{\textbf{B}ehavioral} perturbations require changes to robot movement when solving a task compared to the default setting. Some perturbations can encompass multiple categories - \textit{e.g.} changing the manipulated object (bottom middle) is both visual and behavioral.}
    \label{fig:vis_distribution_shifts}
\end{figure}

\mypar{Benchmark design}. In the initial version, we support 7 skills derived from common manipulation tasks in the DROID dataset: picking, putting, pushing, rotating, stacking, opening, and closing. For our benchmark, we define two task sets based on existing episodes from DROID: \textbf{(i)}~\textit{REALM-base} consisting of 8 tasks testing the pick-place skills and \textbf{(ii)}~\textit{REALM-articulated} consisting of 2 tasks testing the open and close skills on articulated cabinet drawers. Examples of tasks are shown in Fig.~\ref{fig:sim_gen_vis}. In this paper, we distinguish between \textit{skills} that are general and agnostic to objects or scenes and \textit{tasks}, which we consider to be instantiations of a skill manipulating a specific object in a specific scene. We also note that REALM is designed to be extensible and modular, with plans to support more skills, tasks, perturbations, and embodiments in the future. 

\mypar{Choice of perturbations}. To assess the robustness of VLAs under variable conditions, we implemented perturbations that change the visual, behavioral, and semantic properties of the tasks and environments. Table \ref{tab:distribution_shifts} lists all perturbations along with their categorization and implementation. We adopt 14 of the 22 perturbations from the $\bigstar$-Gen taxonomy~\cite{gao2025generalist_taxonomy} and introduce a separate \textbf{V}-LIGHT perturbations for scene illumination. An illustration is shown in Fig.~\ref{fig:vis_distribution_shifts}. To generate the semantic perturbations, which change the language instruction, we use an off-the-shelf VLM~\cite{geminiteam2025geminifamilyhighlycapable}.   

\begin{table}[t!]
\vspace{3mm}
\caption{\textbf{Full list of supported perturbations in REALM}, following a standardized taxonomy~\cite{gao2025generalist_taxonomy}.}
\label{tab:distribution_shifts}
\centering
\begin{tabular}{lp{6cm}}
\toprule
\textbf{Perturbation} & \textbf{Description \& Implementation} \\
\midrule
Default & A task setting close to the training data. \\
\midrule
\multicolumn{2}{c}{\cellcolor{V}{\textit{Visual}}} \\
\midrule
\textbf{V}-AUG & Randomize \textit{blur} and \textit{contrast}. \\
\textbf{V}-SC & Randomly spawn \textit{new distractors} in the scene. \\
\textbf{V}-VIEW & Random shifts to external \textit{camera pose}. \\
\textbf{V}-LIGHT & Randomize illumination \textit{color} and \textit{intensity}. \\
\midrule
\multicolumn{2}{c}{\cellcolor{S}{\textit{Semantic}}} \\
\midrule
\textbf{S}-PROP & Reference objects based on their properties. \\
\textbf{S}-LANG & Reference similar verbs and remove articles.  \\
\textbf{S}-MO & Reference spatial relationships in the scene. \\
\textbf{S}-AFF & Reference human needs and use cases. \\
\textbf{S}-INT & Reference facts about the world that typically require knowledge from Internet-scale text data. \\
\midrule
\multicolumn{2}{c}{\cellcolor{B}{\textit{Behavioral}}} \\
\midrule
\textbf{B}-HOBJ & Randomize manipulated object \textit{mass}. \\
\midrule
\multicolumn{2}{c}{\cellcolor{VB}{\textit{Visual+Behavioral}}} \\
\midrule
\textbf{VB}-POSE & Randomize manipulated \textit{object pose}. \\
\textbf{VB}-MOBJ & Randomize object \textit{size} and \textit{shape}. \\
\midrule
\multicolumn{2}{c}{\cellcolor{SB}{\textit{Semantic+Behavioral}}} \\
\midrule
\textbf{SB}-NOUN & Reference \textit{another known object in the scene}.  \\
\textbf{SB}-VRB & Change the \textit{tested skill} for another compatible one. \\
\midrule
\multicolumn{2}{c}{\cellcolor{VSB}{\textit{Visual+Semantic+Behavioral}}} \\
\midrule
\textbf{VSB}-NOBJ & Sample a \textit{new unseen manipulated object}.  \\
\bottomrule
\end{tabular}
\end{table}

\begin{table}[t]
\caption{\textbf{Tiered progression rubric.} Breakdown of (equally weighted) stages for all tested skills in REALM.}
\label{tab:task_progression}
\centering
\begin{tabular}{c c p{6cm}}
\toprule
\textbf{Set} & \textbf{Skill} & \textbf{Tiered task progression} \\
\midrule
\multirow{5}{*}{\rotatebox[origin=c]{90}{\textbf{Base}}} & \textit{Put} & Reach $\rightarrow$ Grasp $\rightarrow$ Lift $\rightarrow$ Move Close $\rightarrow$ IsInside \\
& \textit{Pick} & Reach $\rightarrow$ Grasp $\rightarrow$ Lift \\
& \textit{Stack} & Reach $\rightarrow$ Grasp $\rightarrow$ Lift $\rightarrow$ Move Close $\rightarrow$ IsOnTop \\
& \textit{Push} & Reach $\rightarrow$ Touch $\rightarrow$ IsToggledOn \\
& \textit{Rotate} & Reach $\rightarrow$ Grasp $\rightarrow$ Rotate $45^\circ$ \\
\midrule
\multirow{3}{*}{\rotatebox[origin=c]{90}{\centering\textbf{Articulated}}} & \textit{Open} & Reach $\rightarrow$ Touch \& Move $\rightarrow$ Open 50\% $\rightarrow$ Open 75\% $\rightarrow$ Open 95\% \\
& \textit{Close} & Reach $\rightarrow$ Touch \& Move $\rightarrow$ Closed 50\% $\rightarrow$ Closed 75\% $\rightarrow$ Closed 95\% \\
&  \\
\bottomrule
\end{tabular}
\end{table}

\mypar{Metrics}. To capture more granularity than a binary success rate, we define a tiered progression for each skill as a sequence of discrete states that need to be achieved in specific order as shown in Table~\ref{tab:task_progression}. Progression ranges from zero to one, where zero corresponds to a complete failure (e.g., failing to reach the manipulated object) and one corresponds to achieving all states in the ordered sequence for a certain skill. Each intermediate stage is weighted equally. When designing and evaluating task progression, we follow best practices for empirical evaluation~\cite{kressgazit2024empricalrobotevals}, but the exact technical details are relegated to the benchmark documentation.

\begin{figure}[t!]
    \centering
    \vspace{3mm}
    \includegraphics[width=\columnwidth]{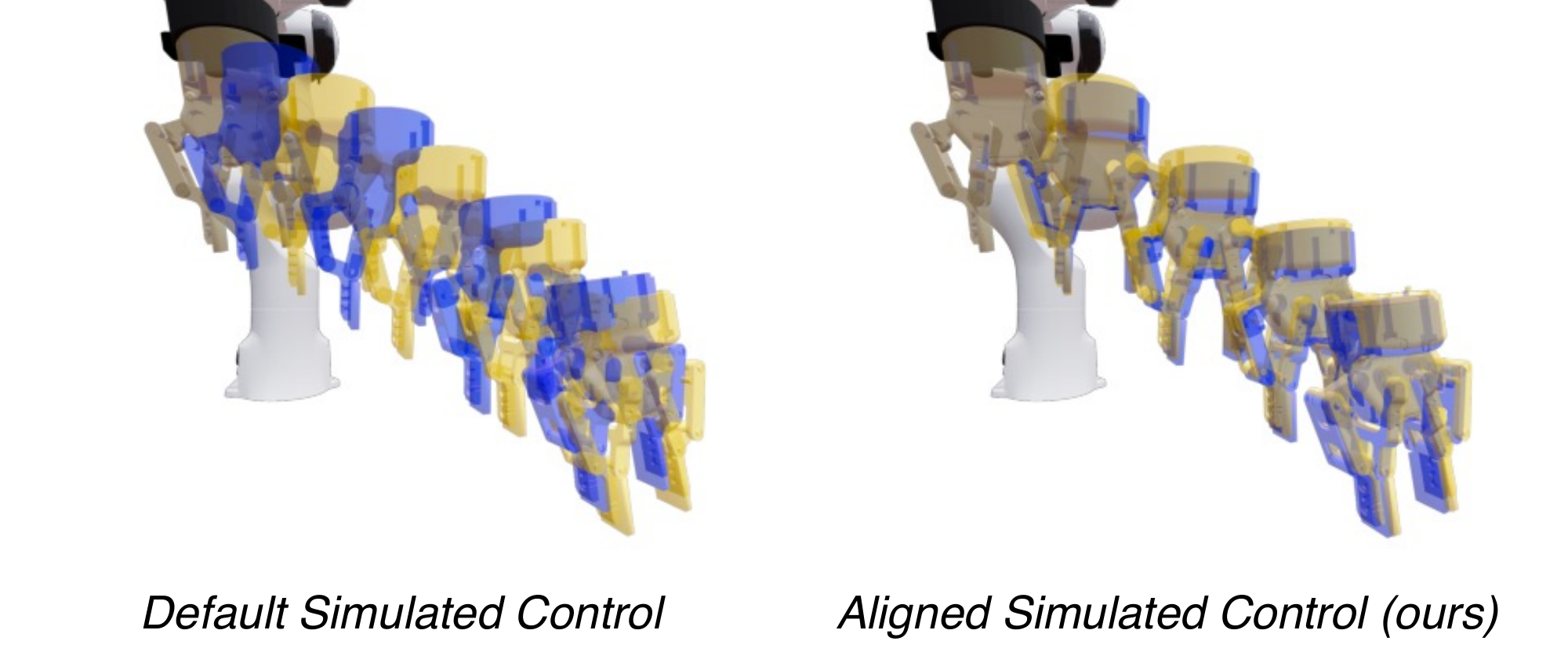}
    \caption{\textbf{Visualization of the aligned control.} We show a trajectory replay in simulation using the default controller (left) and our aligned control (right). \textbf{\color{control_yellow}Yellow} 
    trajectory represents the ground truth from a real robot and \textbf{\color{blue}blue}
    is from simulation. Our system identification results in significantly more realistic trajectory following.}
    \label{fig:controll_alignment}
\end{figure}

\begin{figure}[b!]
    \centering
    \includegraphics[width=\columnwidth]{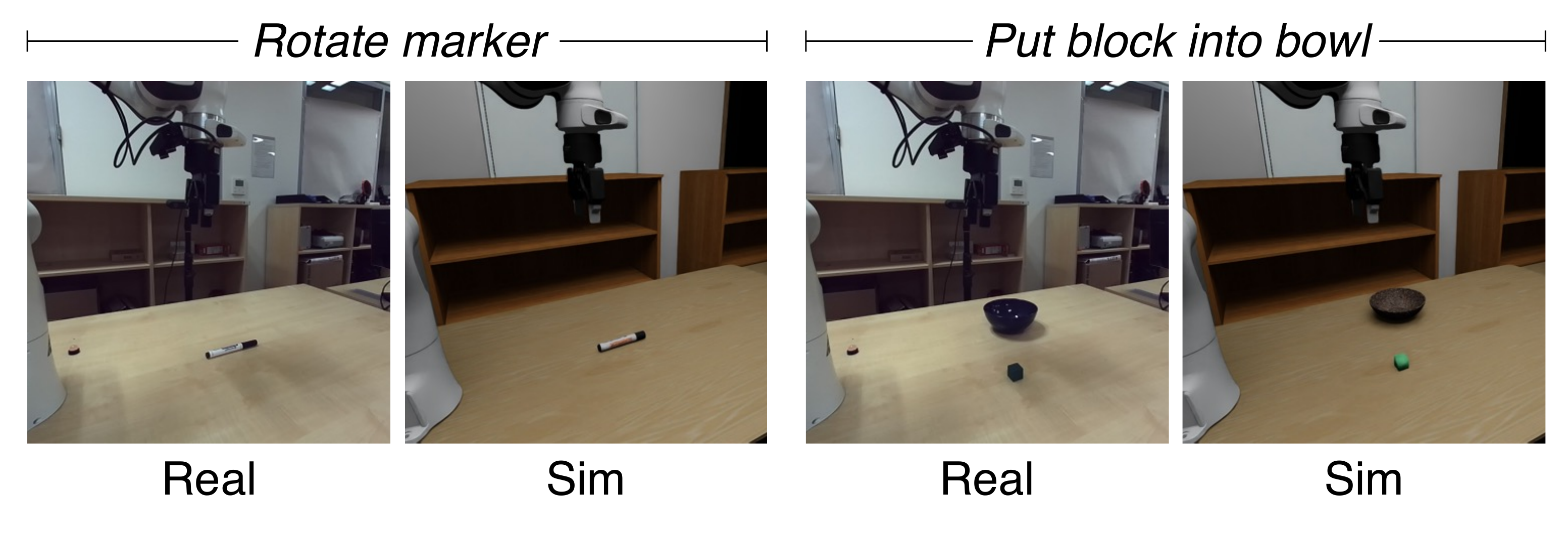}
    \caption{\textbf{Simulation of our set-up}. To measure the real-to-sim gap, we compare the task progression on a real (unseen) set-up in our lab (left) and its digital cousin in sim (right).}
    \label{fig:real2sim_task_overview}
\end{figure}

\begin{figure*}[t]
    \centering
    \vspace{3mm}
    \includegraphics[width=\textwidth]{figures/corr_plot_perturbations.pdf}
    \caption{\textbf{Sim-to-real validation of REALM}. Task progression is shown in the real-world (x-axis) and simulation (y-axis). \textbf{Left}: We show a strong Pearson correlation ($r$) with datapoints close to identity (gray dashed line) and a low Mean Maximum Rank Violation (MMRV) on 7 tasks under 5 visual and behavioral perturbations. \textbf{Right}: The results are also highly correlated under individual perturbations. We observe a p-value of $p < 0.001$ for the correlation between real and simulated rollouts for all settings indicating that REALM is a strong proxy for real-world performance.}
    \label{fig:real2sim_correlation}
    \vspace*{-4mm}
\end{figure*}

\mypar{Choice of embodiment.} Due to the relatively high complexity of precisely aligning the robot control between simulation and reality, we selected DROID~\cite{khazatsky2025droid} as our sole target embodiment. DROID is currently one of the largest available open-source datasets for robot learning and the platform is widely used, including in the crowd-sourced real-world RoboArena~\cite{atreya2025roboarena} evaluation stations. 

\mypar{System identification}. To improve realism, we aim to mitigate the control gap between executing a sequence of actions on the real and simulated robot arms as shown in Fig.~\ref{fig:controll_alignment}. In our approach, the underlying controller is re-implemented and parametrized according to the real DROID platform, leaving 14 free parameters to learn for the simulated joint friction ($\bm{\theta}_{\textrm{friction}}$) and armature ($\bm{\theta}_{\textrm{armature}}$). In IsaacSim~\cite{nvidia2025isaaclabgpuacceleratedsimulation} - the framework, which powers our simulation - friction models the mechanical resistance to motion in a joint and armature represents the reflected inertia of the joint motors, which helps with simulation stability. During optimization, we take a set of $N$ real-world and simulated trajectory replay pairs in joint position space $\mathcal{D} =\{\{(\bm{q}_{i,t}^\textrm{real}, \bm{q}_{i,t}^\textrm{sim}); \bm{q}_{i,t} \in \mathbb{R}^7\}_{t=1}^{T}\}^{N}_{i=1}$, where $t$ is the timestep in an episode, $i$ is the index of an episode pair and $\bm{q}_{i,t}$ is a 7D joint angle vector. We minimize the \textit{control alignment loss} over the dataset~$\mathcal{D}$:
\begin{align}
    \mathcal{L}(\bm{\theta}_{\textrm{friction}}, \bm{\theta}_{\textrm{armature}}) = \sum_{i=1}^{N} \sum_{t=1}^{T} \|\bm{q}^{\textrm{real}}_{i,t} - \bm{q}^{\textrm{sim}}_{i,t}\|_2^2 \, .
    \label{eq:control_alignment_loss}
\end{align}
Frequent replaying of the trajectories in simulation during the optimization process is computationally expensive without a heterogeneous parallel simulation capability, which is not readily supported in the version of the underlying simulator we used. As a result, we use a smaller set of $N=3$ trajectories and optimize the specified parameters using the CMA-ES~\cite{uchida2024cmaessafeoptimization} evolutionary algorithm to obtain initial estimates, then perform parameter search with annealing values.
\section{Real-to-sim validation of the simulation}\label{section:real2sim}
The main goal of REALM is to provide reliable insights into VLA generalization under various perturbations, adequately reflecting their real-world performance. In this section, we show results from an extensive empirical real-to-sim validation on 3 state-of-the-art VLAs, 7 tasks, and 5 perturbations totaling nearly 800 pairs of rollouts. 

\begin{figure}[b!]
    \centering
    \includegraphics[width=\columnwidth]{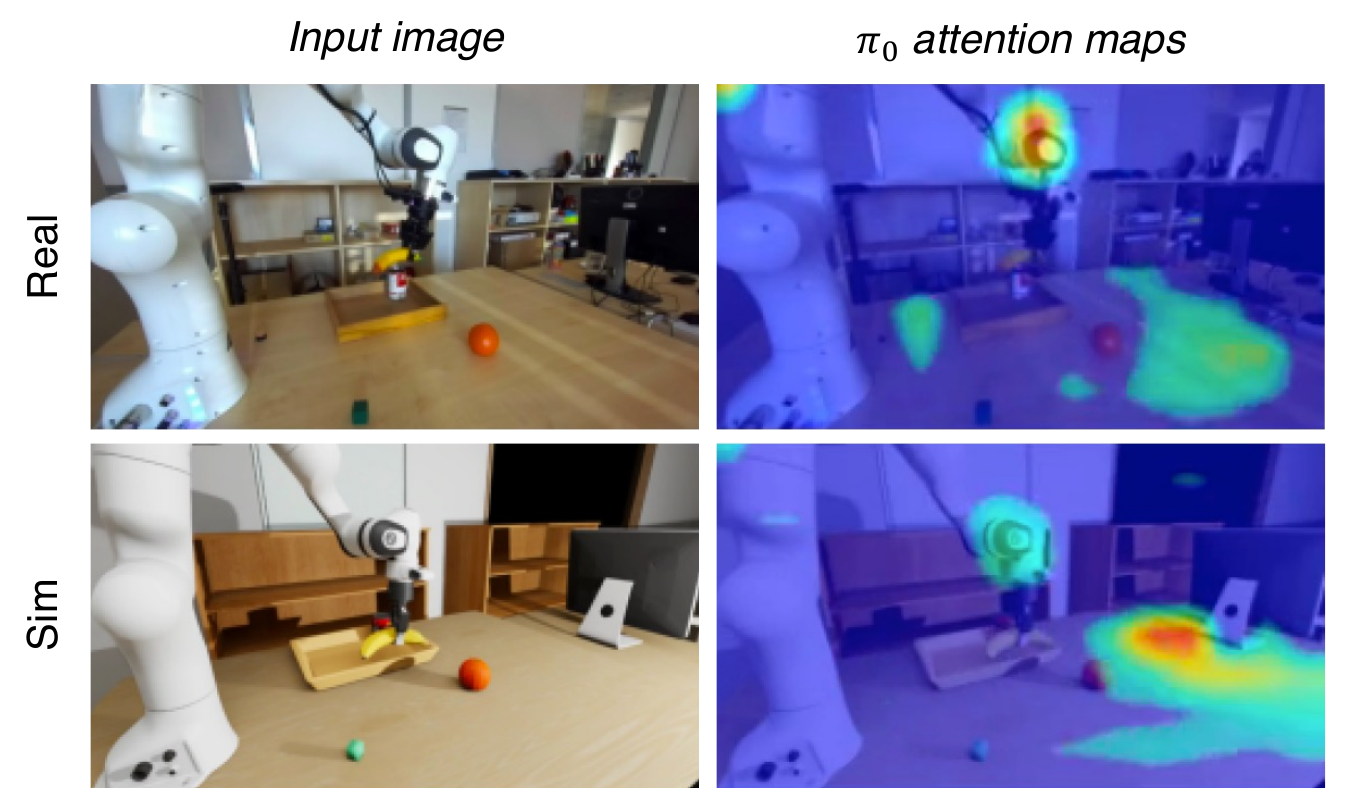}
    \caption{\textbf{Attention maps from the $\pi_0$ action expert}. We replay the same robot trajectory solving a task in reality (top) and simulation (bottom) and observe that the model largely attends to similar patches. We compute the \textit{cosine similarity} between the attention maps from real and simulated images averaged over approx. 280 frames in the video and all layers and attention heads 
    in the $\pi_0$ action expert during the last step of the flow matching process,
    yielding a high similarity score of \textbf{0.85} out of 1.}
    \label{fig:attention_maps}
\end{figure}

\mypar{Validation set-up}. We compare the task progression on a real-world DROID platform and its simulated ``digital cousin"~\cite{dai2024acdc} (shown in Fig.~\ref{fig:real2sim_task_overview}) to validate that our benchmark accurately reflects real-world policy performance. We use three metrics: \textbf{(i)} the Pearson correlation coefficient ($r$) between real and simulated task progression, where higher values indicate better alignment, \textbf{(ii)} a p-value from a standard two-sided t-test on the observed data, where lower values are preferred, and \textbf{(iii)} the Mean Maximum Rank Violation (MMRV) proposed in SIMPLER \cite{li24simpler} to assess the consistency of policy rankings between real and simulated set-ups, where lower values are preferred. In MMRV, a \emph{rank violation}
occurs when one policy outperforms another in simulation, but not in the real-world; the metric then averages over the maximum pair-wise rank violations. In Fig.~\ref{fig:real2sim_correlation}, we show that correlation is consistently high, both overall and under individual visual and behavioral perturbations, with a linear trend close to the identity line and largely consistent relative policy rankings (as shown by the low MMRV).
Lastly, we note that all results shown in Fig.~\ref{fig:real2sim_correlation} are from a digital cousin of a previously unseen scene, suggesting that REALM offers a reliable framework for evaluating generalization in VLA models at scale.

\mypar{Further validation of the visual gap.} Unlike default robot control, we consider the visual fidelity of the simulation to be sufficiently high to not require additional adjustments, such as texture matching~\cite{li24simpler}. We have first shown this by the high correlation between simulated and real-world task progression in Fig.~\ref{fig:real2sim_correlation}. Next, we isolate and quantify the visual gap alone by computing the cosine similarity between attention maps from the $\pi_0$ model's action expert on simulated and real video frames (see Fig.~\ref{fig:attention_maps}). Cosine similarity of the $\pi_0$ attention maps was previously shown to be a good offline indicator for measuring the out-of-distribution status between episodes~\cite{zha2025guidingdatacollectionfactored}, hence, a high score (of 0.85/1 in our case) indicates that using renders from the simulation alone is not enough to cause the model predictions to drastically differ from real-world.

\begin{figure}[t!]
    \centering
    \vspace{3mm}
    \includegraphics[width=\columnwidth]{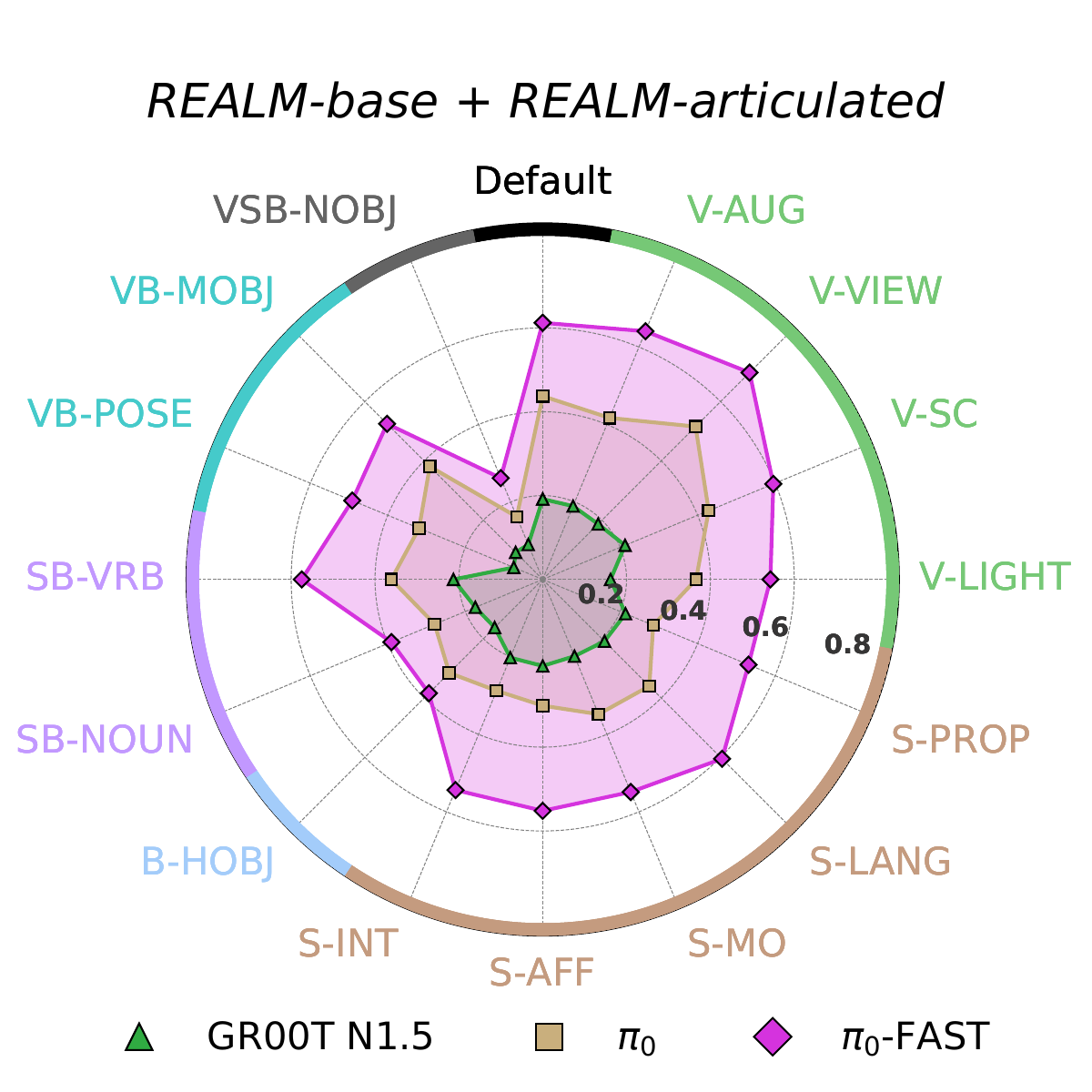}
    \vspace*{-7mm}
    \caption{\textbf{Average task progression on \textit{REALM-base} and \textit{REALM-articulated} tasks}. We show the results on the default setting (black axis) and under each of the 15 perturbations (colored axes). Note that the results are zoomed-in, showing values from 0.0 to 0.8, but the maximum possible value is 1.0. The full non-aggregated results for each of the individual tasks are available on the project website.}
    \label{fig:radar_plot_results_base}
\end{figure}

\section{Results \& Findings}\label{section:Results}
This section presents a detailed evaluation of three VLA models under 15 distinct perturbations on 10 tasks in the \textit{REALM-base} and \textit{REALM-articulated} task sets, resulting in approx. 4,000 simulation rollouts for each model. We discuss these results in the context of four key factors: (\textbf{I}) \textbf{visual generalization}, covering aspects that alter the visual space of the task, (\textbf{II}) \textbf{semantic generalization}, which covers how well the models understand human language and properties of various objects, (\textbf{III}) \textbf{behavioral generalization}, which requires models to adapt the predicted action in order to solve the task, and lastly (\textbf{IV})\textbf{ robustness and task completion}. 

\begin{figure}[t!]
    \centering
    \vspace{3mm}
    \includegraphics[width=\columnwidth]{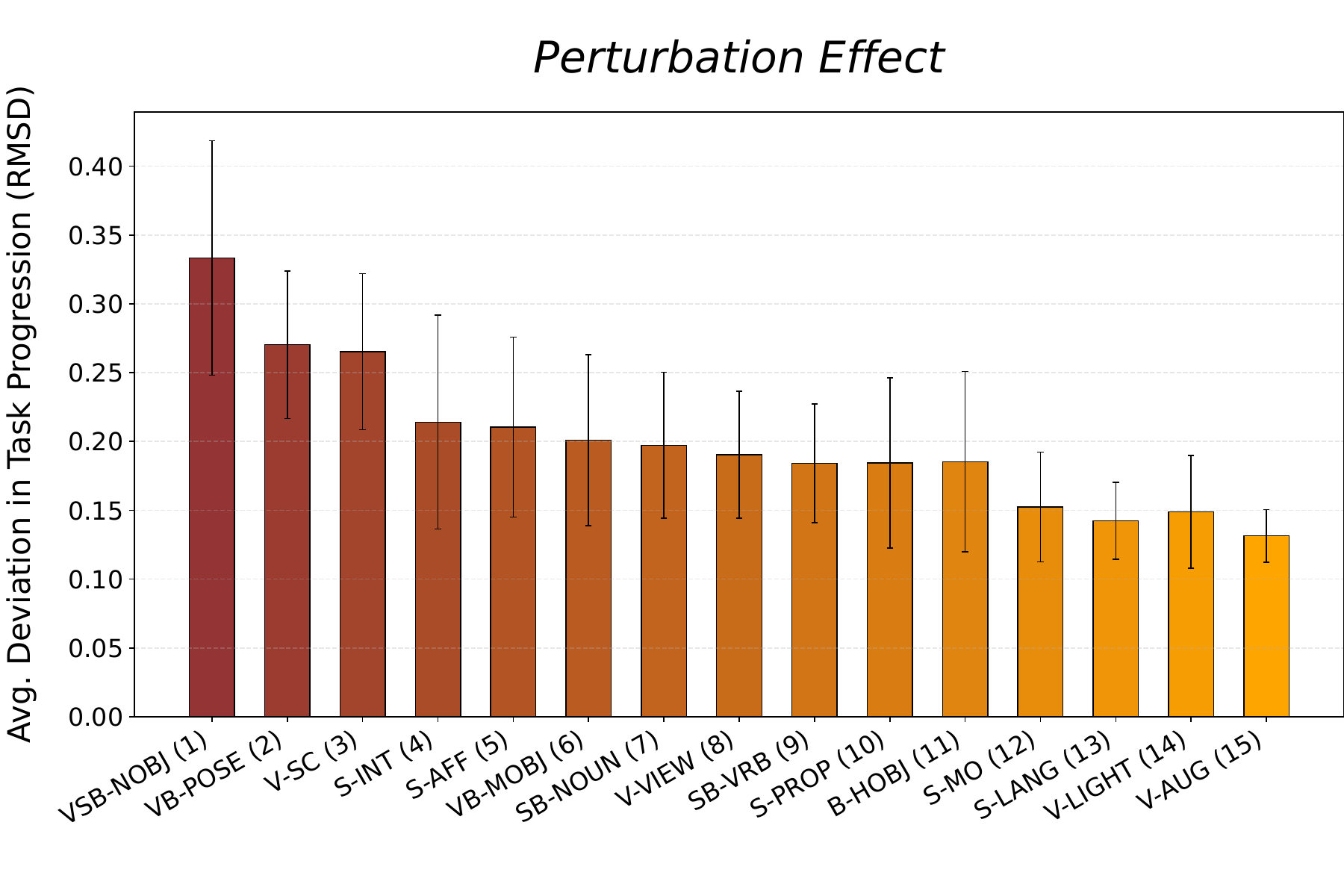}
    \vspace*{-8mm}
    \caption{\textbf{Perturbation effect on task progression.} We show the RMSD between task progression on the default (nominal) and perturbed settings based on data from all 3 models and 10 tasks. \textbf{Ranking:} the perturbations are ranked from left to right based on their \textit{Normalized RMSD} (per model). \textbf{Error bars}: the error bars reflect the deviation in the effects, where tighter bars indicate consistent effect across models and tasks. Conversely, wide bars indicate the effect differs.}
    \label{fig:rmse_plot}
\end{figure}

\begin{figure}[t!]
    \centering
    \vspace{3mm}
    \includegraphics[width=\columnwidth]{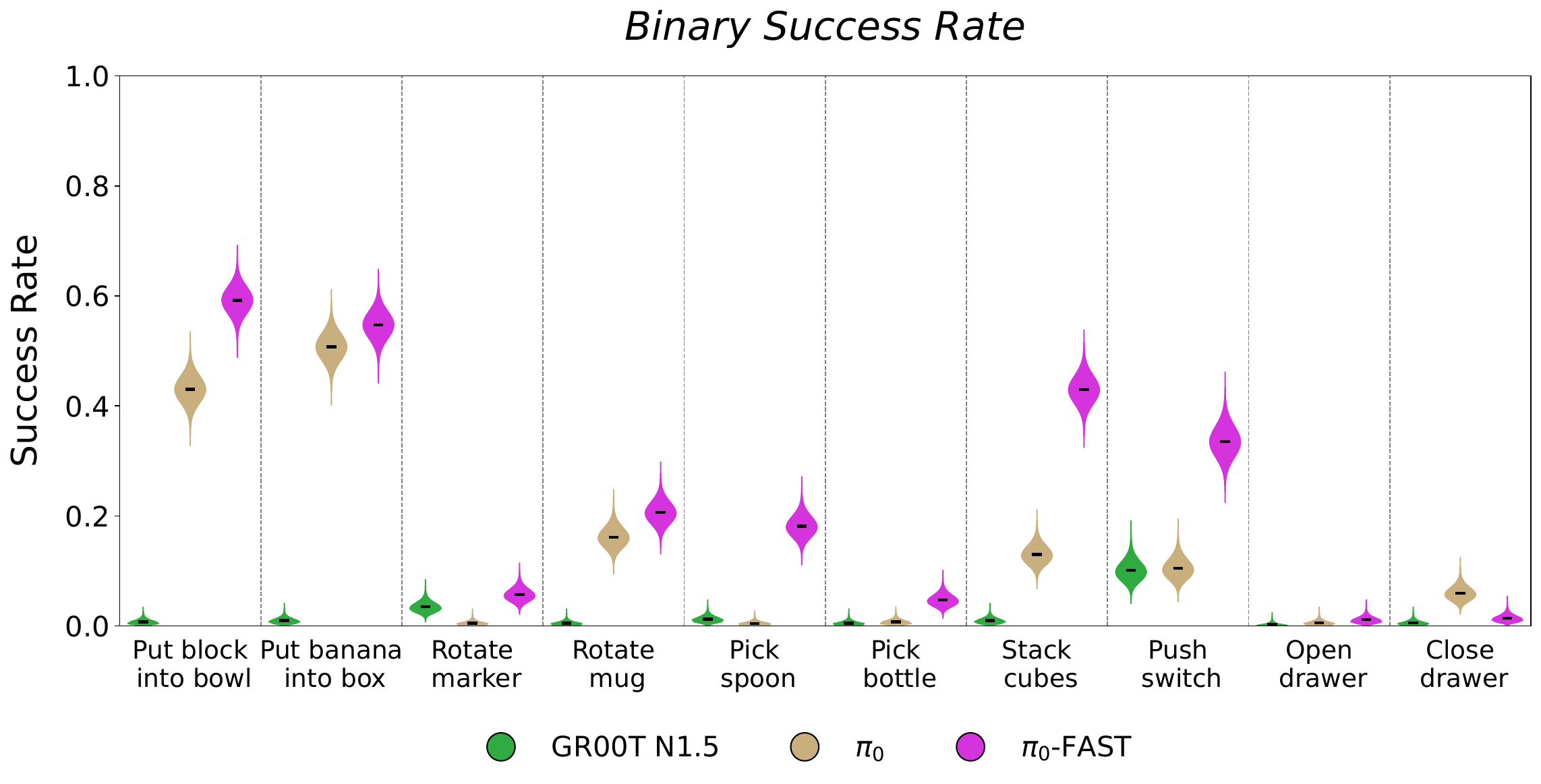}
    \caption{\textbf{Success Rate (y-axis) on the \textit{REALM-base} and \textit{REALM-articulated} task sets (x-axis)}. The Violin plots show Bayesian posteriors of success rates under a uniform Beta prior and the observed data~\cite{trilbmteam2025carefulexaminationlargebehavior}.}
    \label{fig:results_binary_sr}
\end{figure}

\mypar{Evaluation methodology.} In \textbf{Factors I-III}, we primarily discuss the effect of relevant perturbation factors, considering two key metrics: \textbf{(1)} the change in absolute task progression of individual models (see Fig.~\ref{fig:radar_plot_results_base}) and \textbf{(2)} the Root Mean Squared Deviation (RMSD) in task progression (defined in Eq.~\ref{eq:rmsd}), which tells us the overall magnitude of the perturbation effect (see Fig.~\ref{fig:rmse_plot}). \textbf{Factor IV} then focuses on the binary success rate on individual tasks (see Fig.~\ref{fig:results_binary_sr}) and the time to completion across tasks. 

We use the RMSD to quantify the magnitude of the perturbation effect on task progression over all models and tasks, ignoring whether the effect is positive or negative. The metric is computed as follows: 
\begin{align}
    \text{RMSD}(p) = \sqrt{\frac{1}{MT}\sum_{m=1}^{M}\sum_{t=1}^{T}(r_{m,t}^{p}-r_{m,t}^\text{default})^{2}},
    \label{eq:rmsd}
\end{align}
where $p$ is the perturbation whose effect is being measured, $M$ is the number of tested models ($M$=3), $T$ is the number of tasks ($T$=10), and $r$ is the average task progression for a particular model $m$ on task $t$ under active perturbation $p$. The task progression $r$ ranges between 0.0 and 1.0 and the value for each unique $m,t,p$ triplet is computed across a fixed sample size of 25 rollouts, where each individual rollout is scored based on the rubric presented in Table~\ref{tab:task_progression}. The RMSD measures the overall deviation in task progression for perturbation $p$ compared to the ``default" setting. The RMSD value ranges between 0.0 and 1.0. For a truly generalist model, the RMSD would be close to 0.0 for all perturbations.

\mypar{Evaluated models}. We evaluate three state-of-the-art VLA models: $\pi_0$~\cite{black2024pi0}, $\pi_0$-FAST~\cite{pertsch2025fastefficientactiontokenization}, and GR00T N1.5~\cite{nvidia2025gr00tn1openfoundation}.
The $\pi$ models are evaluated using existing open-source checkpoints already fine-tuned on DROID, while we fine-tune GR00T ourselves to operate in the same action space as the $\pi$ models.
We observe that $\pi_0$-FAST achieves the highest overall task progression across all perturbations, clearly outperforming other models (see the magenta curve in Fig.~\ref{fig:radar_plot_results_base}). In contrast, GR00T (green curve) exhibits substantially lower task progression, which makes the measured effects of individual perturbations less informative. As a result, when discussing the generalization capabilities of individual models, we focus solely on $\pi_0$ and $\pi_0$-FAST.

\mypar{Factor I: visual generalization}. 
Analysis of the purely visual perturbations (V-AUG, V-VIEW, V-SC, V-LIGHT) reveals a consistent and noticeable impact on model performance. This is quantitatively shown in Fig.~\ref{fig:rmse_plot}, where all these factors exhibit an average task progression RMSD of at least 0.12. Correspondingly, both $\pi$ models also display visible shifts (both positive and negative) in absolute task progression across these perturbed settings (see Fig.~\ref{fig:radar_plot_results_base}). Examining the individual factors, we see that Blur \& Contrast (15 V-AUG) and Scene Illumination (14 V-LIGHT) rank as the least impactful perturbations (see Fig.~\ref{fig:rmse_plot}). Although the models are not entirely robust to the visual perturbations, we hypothesize their lower severity is a consequence of the high diversity in visual conditions inside the DROID training data, as well as the use of pre-trained VLM backbones. In contrast, more visually-altering perturbations such as viewpoint change (8 V-VIEW) and distractors (3 V-SC) have a substantially stronger effect (see Fig.~\ref{fig:rmse_plot}). In specific instances, these visual perturbations - particularly V-VIEW - result in improvements to the absolute task progression for both $\pi_{0}$ and $\pi_{0}$-FAST when compared to the default setting (see Fig.~\ref{fig:radar_plot_results_base}). While the $\pi$ models are clearly not robust enough to visual perturbations overall, these observed increases in absolute performance indicate that the models can maintain some of their capabilities across a variety of settings.

\mypar{Factor II: semantic generalization}. 
The semantic perturbations (S-PROP, S-LANG, S-MO, S-AFF, and S-INT) are surprisingly challenging for both $\pi$ models, despite large-scale pre-training of their VLM backbones. We observe that $\pi_{0}$ struggles disproportionately more in comparison to $\pi_{0}$-FAST (see the brown curve in Fig.~\ref{fig:radar_plot_results_base}). This significant difference suggests that the diffusion training objective of $\pi_{0}$ harms the language understanding capabilities. The large drop in task progression observed for $\pi_{0}$ also appears to largely contribute to an increased impact (in terms of large RMSD) across all semantic perturbations seen in Fig.~\ref{fig:rmse_plot}. As a direct consequence, perturbations that require knowledge of the world (4 S-INT and 5 S-AFF) have the most profound effect. Lastly, understanding spatial relationships (12 S-MO) has a surprisingly low effect on model performance, despite this information generally not being explicitly represented in the robotics datasets, including DROID.

\mypar{Factor III: behavioral generalization}.
This section focuses on behavioral perturbations (B-HOBJ, SB-NOUN, SB-VRB, VB-POSE, VB-MOBJ, and VSB-NOBJ). These factors require adaptation of motor control and often cover additional semantic and visual changes as well. This makes the set-up inherently challenging, leading to large drops in absolute performance (see Fig.~\ref{fig:radar_plot_results_base}) and a high effect (in terms of RMSD) on model performance overall (see Fig.~\ref{fig:rmse_plot}). Examining the individual perturbations, we observe that both $\pi$ models generalize relatively well across different skills on the same manipulated object (9 SB-VRB) with smaller performance drops. Conversely, they struggle to translate skills consistently across multiple objects (SB-NOUN and VSB-NOBJ) with significant drops in absolute task progression (and large RMSD). We also see a stark difference between objects that are an existing part of the scene and well represented in the DROID dataset (7 SB-NOUN) and unseen objects (1 VSB-NOBJ), where the latter effects both $\pi$ models considerably more (see Fig.~\ref{fig:rmse_plot}). This result is consistent with our expectation that behavioral adaptation, especially to unseen objects, is highly challenging for current VLAs. Changing properties of the manipulated object, such as its shape (6 VB-MOBJ) or mass (11 B-HOBJ), also negatively effects both $\pi$ models, but to a lesser extent (see Fig.~\ref{fig:rmse_plot}). Finally, changing the pose of the manipulated object (2 VB-POSE) has a significant effect on both $\pi_0$ and $\pi_0$-FAST, with the perturbations effect ranking 2nd overall and performance dropping by 0.12 for both models (see Fig.~\ref{fig:radar_plot_results_base}).

\mypar{Factor IV: robustness and task completion}.
As shown in Fig.~\ref{fig:results_binary_sr}, $\pi_0$-FAST tends to have the highest binary success rate on 9 out of 10 tasks, with $\pi_0$ showing better or comparable performance on only 4 tasks. This result is consistent with $\pi_0$-FAST showing the highest average task progression on both the default setting and under all 15 perturbation factors (see Fig.~\ref{fig:radar_plot_results_base}). We also observe that both $\pi$ models take comparably long to complete the tasks, averaging to around 20 seconds (see supplementary video). GR00T generally takes around 30 seconds with a significant variance in the resulting times. In all cases, the solutions take long given the simple manipulation tasks, which indicates that the models are still struggling in unseen environments.

\section{Conclusions and Lessons Learned}
In this work, we introduce REALM: a new high-fidelity simulation environment and benchmark for generalization in robotic manipulation, and perform a real-to-sim validation showing that our results can serve as a proxy for real-world performance. We then evaluate three VLAs under 15 perturbation factors and show that generalization and robustness are still far from solved for these models.

Based on the conducted experiments, we draw the following conclusions: \textbf{(i)} High-fidelity simulation with aligned robot control can serve as a valuable proxy for real-world performance and allow for more informative benchmarks. \textbf{(ii)} Despite using VLM backbones pre-trained on Internet-scale data, there is a noticeable drop in performance from purely semantic perturbations. \textbf{(iii)} There is still a noticeable sensitivity to camera view for all models despite the unusually high diversity of viewpoints in the DROID dataset. \textbf{(iv)} Behavioral generalization across objects and their properties is the most challenging for all tested models. \textbf{(v)} Conversely, all tested models seem to generalize well across known skills when the manipulated object remains the same. \textbf{(vi)} Reliability and robustness, especially under perturbations, are still highly challenging, and models exhibit relatively low overall success rates.

While we recognize the tremendous progress that enabled VLAs to start performing manipulation tasks in unseen settings, including many scenes in our simulation, our results indicate that current models still lack the capabilities for autonomous real-world deployment. We also hypothesize that the full fine-tuning on DROID data used for all three models harms the generalization capabilities of the VLM backbones, leading to many of the observed failures. 

\mypar{Limitations}. One limitation for benchmarking generalization is that some models achieve low performance on certain tasks, which typically makes most perturbations uninformative. We mitigate this issue by designing the default task settings to be close to in-distribution for the DROID dataset. Additionally, while we believe that DROID is a widely-available platform and its simulation will be useful to the robot learning community, we also hope to support more robots in the future. Despite these limitations, we believe this work opens up an exciting possibility of systematically studying generalization in VLAs.

\mypar{Acknowledgments}.
This work was supported by the European Union’s Horizon Europe projects AGIMUS (No. 101070165), euROBIN (No. 101070596), ERC FRONTIER (No. 101097822), and ELLIOT (No. 101214398). Pavlo Yefanov (PY) and Georgy Ponimatkin (GP) were also partly supported by Grant Agency of the Czech Technical University in Prague under allocations SGS25/158/OHK3/3T/13 (PY) and SGS25/156/OHK3/3T/13 (GP). Martin Sedlacek was partly supported by the ELLIS Unit Amsterdam as part of the MSc Honours Programme. Compute resources and infrastructure were supported by the Ministry of Education, Youth and Sports of the Czech Republic through the e-INFRA CZ (ID:90254) and by the European Union’s Horizon Europe project CLARA (No. 101136607).

\bibliographystyle{IEEEtran}
\bibliography{bibliography}

\addtolength{\textheight}{-12cm}

\end{document}